\documentclass[10pt,twocolumn,letterpaper]{article}

\usepackage{cvpr}              
\usepackage{graphicx}
\usepackage{amsmath}
\usepackage{amssymb}
\usepackage{booktabs}
\usepackage{bbding}
\usepackage{amssymb}
\usepackage{multirow}
\usepackage{makecell}

\definecolor{cvprblue}{rgb}{0.21,0.49,0.74}
\usepackage[pagebackref,breaklinks,colorlinks,allcolors=cvprblue]{hyperref}

\def\paperID{11144} 
\def\confName{CVPR}
\def\confYear{2025}

\title{DiN: Diffusion Model for Robust Medical VQA with Semantic Noisy Labels}

\author{Erjian Guo\textsuperscript{\rm 1}~~~~Zhen Zhao\textsuperscript{\rm 2}~~~Zicheng Wang\textsuperscript{\rm 1}~~~~Tong Chen\textsuperscript{\rm 1}~~~~Yunyi Liu\textsuperscript{\rm 1}~~~~Luping Zhou\textsuperscript{\rm 1}\thanks{Corresponding author, luping.zhou@sydney.edu.au. Zhou is supported by Australian Research Council (ARC DP200103223).} \\
\textsuperscript{\rm 1}University of Sydney\hspace{16mm}
\textsuperscript{\rm 2}Shanghai AI Laboratory\hspace{16mm} \\
}

\begin{document}
\maketitle
\begin{abstract}
Medical Visual Question Answering (Med-VQA) systems benefit the interpretation of medical images containing critical clinical information. However, the challenge of noisy labels and limited high-quality datasets remains underexplored. To address this, we establish the first benchmark for noisy labels in Med-VQA by simulating human mislabeling with semantically designed noise types.  More importantly, we introduce the DiN framework, which leverages a diffusion model to handle noisy labels in Med-VQA. Unlike the dominant classification-based VQA approaches that directly predict answers, our Answer Diffuser (AD) module employs a coarse-to-fine process, refining answer candidates with a diffusion model for improved accuracy. The Answer Condition Generator (ACG) further enhances this process by generating
task-specific conditional information via integrating answer embeddings with fused image-question features. To address label noise, our Noisy Label Refinement(NLR) module introduces a robust loss function and dynamic answer adjustment to further boost the performance of the AD module. Our DiN framework consistently outperforms existing methods across multiple benchmarks with varying noise levels~\footnote{Code: \url{https://github.com/Erjian96/DiN}}.
\end{abstract}    
\section{Introduction}
\label{sec:intro}

\begin{figure}[h]
\centering
\includegraphics[width=1\linewidth]{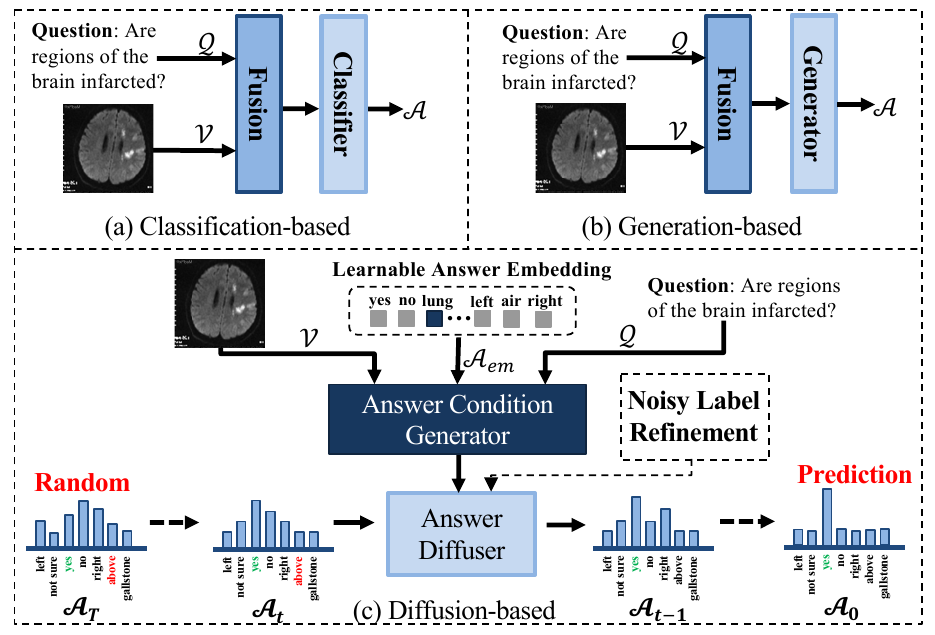}
\caption{\textbf{Comparison between our diffusion-based method and previous Med-VQA approaches.} Previous Med-VQA methods, designed for clean label datasets, include classification-based (a) and generation-based (b) approaches. In contrast, our diffusion-based method (c) classifies answers from a generative perspective and denoises answers by Answer Diffuser with the support of Answer Condition Generator and the Noisy Label Refinement module (employed only during the training process).}
\label{diffusion}
\end{figure}

Medical Visual Question Answering (Med-VQA) tasks focus on accurately responding to medical inquiries by integrating visual information from medical images~\cite{lin2023medical,demirhan2023survey,wang2022medical}. It holds great potential in applications like assisting diagnosis, enhancing patient understanding, and supporting clinical education~\cite{bian2024diffusion,shen2017deep}. 
Recently, VQA models have achieved impressive results in the natural VQA domain~\cite{song2022pixels,he2021towards,yu2020cross}. This success has largely been driven by the availability of large-scale natural VQA datasets with well-structured accurate annotations~\cite{lu2023multi,lin2023fine}. However, for Med-VQA, the annotation process poses unique challenges. It requires specialized medical expertise and skills from annotators, leading to increased complexity and higher potential for label noise~\cite{seenivasan2022surgical,sharma2021medfusenet,zhao2024alternate}. Additionally, inter-observer variability among medical professionals further complicates the annotation process, contributing to labeling inconsistencies~\cite{hu2024interpretable,ren2020cgmvqa,wu2024hallucination}. While the issue of noisy labels is a well-known challenge in machine learning, it has not been thoroughly explored in Med-VQA, despite the need for highly accurate medical annotations.

\begin{figure*}[h]
\centering
\includegraphics[width=0.95\linewidth]{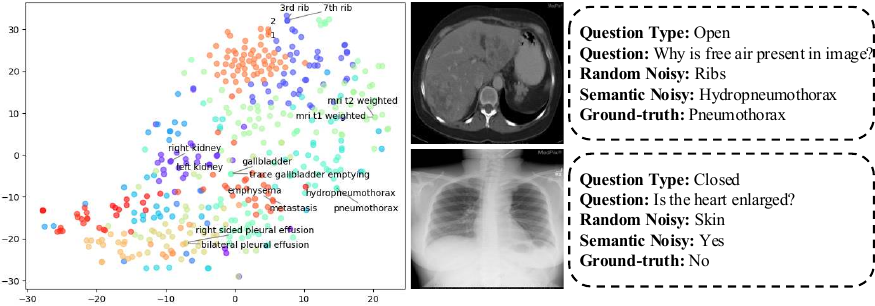}
\caption{\textbf{Visualization of the semantic answer pairs in t-SNE and examples of noisy answers.} Similar colors represent the degree of feature similarity from a pretrained BERT model. We randomly display 8 semantic pairs in the t-SNE plots of VQA-RAD dataset. Ground-truth labels are replaced with their semantic pairs, which we refer to as semantically noisy answers~\cite{zhang2023learning}. }
\label{show_noise}
\end{figure*}

While existing label-denoising methods have shown promise, they are primarily designed for uni-modal tasks like image classification or segmentation \cite{wei2021learning,li2022selective,zhao2023augmentation}.  Directly applying them for the multi-modal Med-VQA task is not suitable for two key reasons. 
First, conventional noise models, such as random label shuffling, fail to capture the semantic nuances essential in Med-VQA, where label noise often retains medically relevant information rather than arbitrary noise.
We argue that noisy labels in VQA should retain semantic coherence aligned with human cognitive levels. 
As shown in \cref{show_noise}, examples of noisy QA pairs and their corresponding t-SNE representations illustrate this difference: unlike the arbitrary and meaningless answers generated by random noise, our constructed semantic noise resembles real-world scenarios, closer to human mislabeling.
Second, these methods are not equipped to handle the complexity of Med-VQA, which integrates both image and text modalities. 
For example, co-teaching methods~\cite{xia2023combating} add considerable model complexity by requiring two branch networks, while Noise Transition Matrix (NTM) methods~\cite{guo2023handling} struggle with noisy label distribution estimation as the number of Med-VQA classes grows. 
While current leading methods like SNLC~\cite{zhang2023learning} in the natural VQA domain focus on improving representation through robust contrastive loss, these studies address noise in a relatively coarse way and fall short when dealing with the medically relevant label noise in Med-VQA. For instance, with a 20\% noise ratio, SNLC~\cite{zhang2023learning} achieves 62\% accuracy on Med-VQA with random noise but drops to 58\% when handling our semantic noise.
In summary, our constructed semantic noise in Med-VQA is both practically relevant and presents a more challenging task to effectively handle.

In the literature, two main approaches have emerged for medical VQA: classification-based and generation-based methods, as shown in \Cref{diffusion}. Classification-based methods are generally straightforward and efficient but often struggle with low-frequency answers, which becomes especially pronounced in complex domains like Med-VQA~\cite{eslami2023pubmedclip,ambati2018sequence}. 
In contrast, generation-based methods offer more flexibility by constructing answers instead of selecting from predefined categories; however, they inevitably generate unrealistic or non-existent answers, reducing overall accuracy and reliability~\cite{nguyen2019overcoming,liu2023q2atransformer}.
To this end, motivated by the success of diffusion models in label refinement~\cite{betker2023improving, saharia2022photorealistic}, we propose leveraging them to bridge the gap between classification and generation approaches, enabling answer classification from a generative perspective. 
Diffusion models can offer a balance between classification and generative approaches for Med-VQA. Prior studies \cite{han2022card, chen2024label} have also demonstrated that the probabilistic nature of diffusion models effectively handles data and label uncertainty, leading to more robust predictions. 
Conditioned on discriminative feature and label information, diffusion models offer a promising solution for the noisy labels in Med-VQA, using their stochastic nature to reconstruct class prototypes during reverse sampling and generate probability intervals across class labels.

To improve Med-VQA performance in the presence of noisy labels and harness the benefits of diffusion models, we propose the conditional Diffusion model for Med-VQA with Noisy labels (DiN) framework to address challenging semantic noisy labels. 
To mimic human-like mislabeling,  we construct the training set by replacing ground-truth answers with semantically related noisy pairs, selected using a pre-trained BERT model.  
To the best of our knowledge, this is the first study addressing Med-VQA with noisy labels and the first application of diffusion models to this task. 
Our framework comprises three key modules. The Answer Denoiser (AD) module leverages robust conditional diffusion models, utilizing their powerful gradual refinement capabilities for Med-VQA. 
As shown in Figure~\ref{diffusion}, the generative process begins with a noisy label estimation, which is progressively refined to recover a clean label, analogous to the reverse denoising process in diffusion models. This allows coarse-to-fine filtering of potential answers, significantly improving accuracy and robustness. 
However, the effectiveness of the AD module's refinement is highly dependent on the generation conditions, and the lack of accurate clean labels exacerbates this challenge. To this end, we further propose the Answer Condition Generator (ACG) and Noisy Label Refinement (NLR) module to provide more accurate and reliable conditions.
The ACG adopts a transformer decoder to fuse multi-modal image and question features, as well as learnable candidate embedding for delivering more informative generation conditions. The NLR, utilized only during training, refines noisy ground-truth labels dynamically through pseudo-labels generated by an auxiliary classifier to supervise the AD module. This synergy across modules enhances precision and robustness in managing noisy labels for Med-VQA tasks. 

Our main contributions are summarized as follows.
\begin{itemize}    
    \item  We propose the DiN framework, the first to address Med-VQA tasks with noisy labels (NM-VQA) by integrating a diffusion model that classifies answers from a generative and coarse-to-fine perspective. 

    \item We introduce semantic noise types tailored to Med-VQA tasks and construct challenging benchmark datasets that simulate human labeling errors based on an thorough analysis of existing datasets.

    
    \item We design the Answer Condition Generator (ACG) and Noisy Label Refinement (NLR) modules to support the diffusion-based Answer Diffuser (AD). 
    ACG guides AD with conditions, while NLR refines noisy labels using pseudo-labels from an auxiliary classifier. 

    \item Our method consistently outperforms existing approaches on two Med-VQA datasets with varying noise ratios, demonstrating its robustness and effectiveness.
\end{itemize}

\section{Related Work}
\label{sec:related_work}

\subsection{Medical Visual Question Answering}
 Med-VQA is still in its early stages, and its current performance remains unsatisfactory. Most methods, such as those in~\cite{liu2023q2atransformer,do2021multiple,nguyen2019overcoming,ren2020cgmvqa}, adopt a classification-based approach, where each answer is treated as a class, and a model predicts the answer by applying classification to the fused image-question features. This classification-based approach simplifies the task by reducing its complexity and narrowing the answer search space. Alternatively, some studies~\cite{khare2021mmbert,eslami2021does} approach VQA as a generation problem, generating answers word by word. These generation-based methods face challenges due to the vast search space, often leading to inaccurate answers, which limits their effectiveness in medical VQA. 
However, existing Med-VQA methods assume that the entire training dataset consists of clean labels, overlooking the challenge of noisy labels (NM-VQA). To address this gap, we introduce Med-VQA datasets with noisy labels and propose a solution for the NM-VQA problem.

\subsection{Learning with Noise Labels and Noise Types}
Learning with noise labels has been extensively studied in recent works. For example, some methods estimated noise transition matrices to correct label noise in a class-wise manner~\cite{guo2023handling,li2020dividemix}. Other popular methods reduced the effects of noisy labels by using noise-tolerant loss functions~\cite{zhang2023learning}. However, these methods tend to perform poorly when confronted with high noise rates and a large number of classes in Med-VQA tasks. One typical alternative way is coteaching-based methods CoDis~\cite{xia2023combating}, which require two branch networks, and introduce a double number of additional parameters, thereby exacerbating the issue of model complexity in Med-VQA tasks that necessitate both image and text encoders within a single network. Given these challenges, it is valuable to explore methods for developing robust Med-VQA models under reasonable noisy data.
Most methods to deal with artificially synthesized noisy labels were initially proposed for classification tasks~\cite{karimi2020deep}. In these studies, synthesized noisy labels typically focus on both symmetric and asymmetric label noise. 
Symmetric noise flips labels randomly with equal probability, while asymmetric noise flips labels based on fixed rules. 
However, when it comes to Med-VQA tasks, symmetric and asymmetric noise models appear unreasonable. Unlike the traditionally synthesized label noise for classification, we propose the semantic label noise for Med-VQA tasks.

\subsection{Diffusion Models}
Diffusion models~\cite{betker2023improving, saharia2022photorealistic, DiffSurvey, Diff2, rombach2022high} are a new class of generative model. During training, they initially add noise to labels through a forward process until they approximate a Gaussian distribution. Then, in the reverse process, they learn to denoise and recover the original labels~\cite{DDIM, ho2020denoising, zhang2023adding}. In recent years, their remarkable success in the generative domain, particularly in visual generation~\cite{saharia2022photorealistic}, has been groundbreaking, completely surpassing previous generative models. Due to its progressive refinement of labels and continuous improvements in training and inference efficiency, the diffusion model has been applied in various fields, such as recommendation systems~\cite{wang2023diffusion} and image detection~\cite{chen2023diffusiondet}. For example, the CARD~\cite{han2022card} model transforms deterministic classification into a conditional label generation process, enabling more flexible uncertainty modeling in the labeling process. However, to our knowledge, diffusion models have not yet been applied to tackle Med-VQA tasks. Therefore, to address this gap, we aim to fully leverage the unique characteristics of diffusion models for Med-VQA tasks.
\section{Methods}
\label{sec:methods}

\begin{figure*}[t]
\centering
    \includegraphics[width=0.9\textwidth]{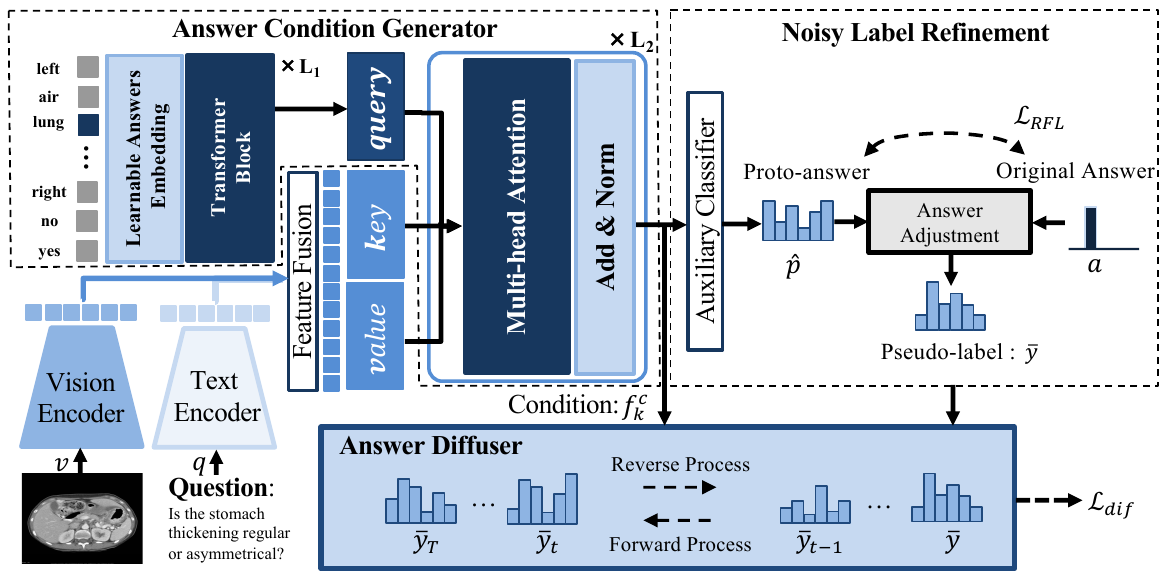}
    \caption{The proposed DiN framework consists of three key modules: 1) Answer Condition Generator (ACG): This module interacts the image and question multi-modal features with features of the image-question pair knowledge from Answer Condition Embedding (ACE) to obtain Med-VQA condition information. 2) Noisy Label Refinement (NLR): This module contain a robust loss function, $\mathcal{L}_{RFL}$, which supervises the proto-answer, to mitigate the impact of noisy original answers on the two encoders' acquisition of medical domain knowledge. The Answer adjustment of NLR uses proto-answers and original noisy anwers to generate pseudo-label to supervise the AD module. 3) Answer Diffuser (AD): This module refines the noisy answer distribution, simulating a generation process to select the correct answers. Notably, we use only the AD Module to predict answers without the NLR module during inference process. }\label{framework_Dif}
\end{figure*}
\label{method}

\subsection{Problem setting}
In the context of Med-VQA tasks with noisy labels (NM-VQA) problem, our task involves an input set of medical images $\mathcal{V}=\{v_i\}_{i=1}^{N}$ and a set of questions $\mathcal{Q}=\{q_j\}_{j=1}^{M}$ with the corresponding answers $\mathcal{A}=\{a_j\}_{j=1}^{M}$. Each image $v_i$ is associated with $M_i$ questions and answers, while $\sum_{i=1}^N M_i$ equals $M$. For the convenience of description, we use the subscript $k$ to represent the index of each image-question-answer pair and there are $K$ image-question-answer pairs in total. Specifically, $v_k \in \mathbb{R}^{H\times W\times C}$ represents an input image with a size of $H\times W$ and $C$ channels, while $a_k \in \left\{ 0, 1\right\}^{L}$ serves as the one-hot encoded ground truth answer label, where $L$ is the total number of answer classes. 
Note that only noisy labels are available during the training stage, while the clean labels are only available during the inference stage and are used solely for validating the model's performance.

\subsection{Overview}
\label{sec:Overview}

The DiN framework, shown in Figure~\ref{framework_Dif}, comprises three main modules: Answer Condition Generator module (ACG) introduced in Sec.~\ref{sec:Answer Condition Generator Module}, Noisy Label Refinement module (NLR) described in Sec.~\ref{sec:Noisy Label Refinement Module}, and Answer Diffuser module (AD) detailed in Sec.~\ref{sec:Answer Diffuser Module}. The input image $v_k$ and question $q_k$ are processed by Vision Encoder and Text Encoder, respectively, to extract visual features $f_k^v$ and question features $f_k^q$. The fused features produce key embeddings $K_{ey}$ and value embeddings $V_{alue}$, which are fed into the ACG module. There, they interact with learnable answer embeddings $Q_{uery}$ to generate an answer-aware condition $f_k^c$. This condition is then used by the AD module to generate the predicted answer $y_k$ via a denoising process. Simultaneously, $f_k^c$ is fed into the NLR module to train an auxiliary classifier, which yields a proto-answer $\hat{y}_k$. The proto-answer is then refined to create a pseudo-label $\bar{y}_n$ used to supervise the generation of the $y_k$. Note that the NLR module will be discarded in the inference stage, and the AD module will generate the answer predictions.

\subsection{Answer Condition Generator Module}
\label{sec:Answer Condition Generator Module}

The newly proposed ACG module is designed to generate an answer-aware condition, $f_k^c$, for Noisy Label Refinement module (NLR) and  Answer Diffuser module (AD). As introduced in Sec.~\ref{sec:Overview}, we obtain visual features $f_k^v$ and question features $f_k^q$ from input images and questions by Vision Encoder and Text Encoder. These features are then fused to produce key embeddings $K_{ey}$ and value embeddings $V_{alue}$, which are fed into the ACG module. In the ACG module, we introduce learnable candidate answer embeddings to
query the existence of each answer class to a given image-question pair. For the $L$ answer classes, we need $L$ candidate answer embeddings initialized randomly and updated during training. A self-attention mechanism is applied to compute relationships between the answer embeddings, yielding query embeddings $Q_{uery}$. Cross-attention is then used to combine $Q_{uery}$ with $K_{ey}$ and $V_{alue}$, producing a fused feature $f_k^f$ that captures the interaction between answer embeddings and image-question features. Finally, the feature $f_k^c = \text{Linear}(f_k^f)$ is generated as conditions to guide the generative process in the AD module, while $f_k^f$ is simultaneously sent to the NLR module to train the auxiliary classifier.

\subsection{Noisy Label Refinement Module}
\label{sec:Noisy Label Refinement Module}
Considering that the model learning will suffer severely from label noise, in this part, we tackle the problem by proposing a Noisy Label Refinement module (NLR), where we first propose a new noise robust loss function to improve the noise resilience of our method and we further come up with an answer adjustment strategy to to generate accurate refined pseudo-label $\bar{y}_k$ to supervise the generation of $y_k$ in the Answer Diffuser module. 

Specifically, we first feed the condition $f_k^c$ into an auxiliary classifier, which is a one-layer linear MLP, to generate the proto-answer probability distribution $\hat{p}_k$. However, on one hand, the noisy supervision $a_k$ will greatly influence the accuracy of the proto-answer. On the other hand, the class imbalance problem in Med-VQA is much more severe than other tasks, such bias will severely influence the answer generation. Therefore, we introduce a new noise robust focal loss function (RFL) to tackle the issues, where we incorporate a symmetric cross entropy loss with a focal loss, aiming to improve noise resilience and balance the effects of class imbalance. The RFL is then formulated as follows:
\begin{equation}
    \begin{aligned}
    \mathcal{L}_{RFL} & = - \sum_{l=1}^L a_k(1-\hat{p}_k)^{\gamma}log \hat{p}_k - \sum_{l=1}^L \hat{p}_k log a_k ,\label{eq_noise}
    \end{aligned}
\end{equation}
where $\gamma$ is the hyper-parameter and we set $\gamma=1$ by default. 

Benefiting from our newly proposed RFL loss, the model is prone to generate correct answers during the training process. Therefore, we further propose an Answer Adjustment (AA) strategy to generate a set of accurate refined pseudo-labels to supervise the generation of $y_k$ in the Answer Diffuser module. In particular, we assume the given answer is clean if the predicted proto-answer matches the original answer, otherwise, we generate a soft label using a weighted combination of the predicted proto-answer $\hat{p}_k$ and the given noisy label distribution $a_k$, which can be denoted as:
\begin{equation}
    \bar{y}_k =
    \begin{cases}
    w_t\hat{p}_j+(1-w_t)a_k,  & \hat{p}_k \neq a_k\\
   a_k, & \hat{p}_k = a_k\\
    \end{cases},\label{new_answer}
\end{equation}
where $w_t$ is the weighting parameter. Note that the learning status can be estimated by the confidence scores of the predictions~\cite{guo2017calibration}, we thereby design $w_t$ based on the model’s prediction reflecting the overall learning status. Given the current training iteration $t$, we denote $w_t$ as:
\begin{equation}
 w_t= \tau w_{t-1}+(1-\tau) \frac{1}{B} \sum_{k=1}^{B} \hat{p}_{k},
\end{equation}
where $\tau \in (0,1)$ is the momentum decay of EMA, set as 0.99. The $B$ indicates the number of samples within a mini-batch.
Through the newly proposed answer adjustment strategy, we can obtain a set of accurate refined pseudo-labels $\bar{y}$ to supervise the generation of the final prediction $y$ in the Answer Diffuser module, which will greatly improve the prediction performance of the model.

\subsection{Answer Diffuser Module}
\label{sec:Answer Diffuser Module}

Given the ability of diffusion-based classifiers to manage data and label uncertainty effectively, we propose a diffusion-based classifier called the Answer Diffuser module. This module addresses the unique challenges of NM-VQA by classifying answers from a generative perspective, which helps mitigate issues like generating non-existent answers—a common problem in handling noisy labels~\cite{han2022card, chen2024label}. By using refined soft labels as supervision, the Answer Diffuser produces accurate, noise-resistant predictions that are particularly robust in noisy-label settings. 

With the condition feature $f^c$ and the refined soft label $\bar{y}$, following the forward and reverse process of the diffusion models, we gradually introduce noise to the refined soft label $\bar{y}$ to obtain the noisy label $\bar{y}_T$ after $T$ steps, and progressively refine a random probability distribution $\bar{y}_T$ to predict the answer distribution $y$ that is close to the pseudo-label distribution $\bar{y}$ after $T$ steps. Notably, the forward can be represented as:
\begin{equation}
    q(\bar{y}_t|\bar{y}_{t-1},f^c) = \mathcal{N}(\bar{y}_t;\mu_t(\bar{y}_{t-1},f^c,t),\beta_t\mathbf{I}),\label{diff}
\end{equation}
\begin{equation}
    \mu_t = \sqrt{1-\beta_t}\bar{y}_{t-1}+(1-\sqrt{1-\beta_t})f^c,
\end{equation}
where $\beta_t$ is the variance schedule. Consequently, obtaining $\bar{y}_t$ with an arbitrary timestep $t$ can be represented as:
\begin{equation}
    q(\bar{y}_t|\bar{y},f^c) = \mathcal{N}(\bar{y}_t;\sqrt{\bar{\alpha}_t}\bar{y}+(1-\sqrt{\bar{\alpha}_t})f^c,(1-\bar{\alpha}_t)\mathbf{I}),
\end{equation}
Where $\alpha_t = 1-\beta_t$ and $\bar{\alpha}_t = \prod_t\alpha_t$.
It should be noted that $\mu_t$ can be viewed as an interpolation between soft psuedo-label $\bar{y}$ from NLR module and the predicted condition $f^c$ from ACG module, which gradually generates task-specific conditional information to suit the NM-VQA task.

The reverse process continuously perform denoising process from $\bar{y}_T$ to obtain $y$, which we hope to be close to $\bar{y}$, where we learn a diffusion model $\theta$ to predict the noise level at each timestep, which can be represented as:
\begin{equation}
    p_{\theta}(\bar{y}_{t-1}|\bar{y}_t, f^c) = \mathcal{N}(\bar{y}_{t-1};\mu_{\theta}(\bar{y}_t,f^c,t), \sigma^2_t\mathbf{I}),
\end{equation}
where the $\mu_{\theta}(\bar{y}_t,f^c,t)$ is predicted mean value of the noise distribution at timestep $t$, $\sigma^2_t$ is $\frac{1-\bar{\alpha}_{t-1}}{1-\bar{\alpha}_{t}}\beta_t$. Our target is to encourage the predicted answer distribution to be close to the refined pseudo-label distribution $\bar{y}$. Therefore, we adopt a MSE loss as a supervsion, which can be denoted as:
\begin{equation}
    \mathcal{L}_{dif} = \text{MSE}(y;\bar{y}).
\end{equation}

\subsection{Training and Inference}
\label{sec:Training and Inference}

During the training process, we jointly optimize $\mathcal{L}_{dif}$ and $\mathcal{L}_{RFL}$, which on one hand targets at generating refined correct pseudo-labels that are robust to label noise as the supervision for our Answer Diffuser, on the other hand, we encourage the Answer Diffuser to generate existent and accurate answer predictions, further reducing the impact of noisy labels. Therefore, the overall loss during training can be expressed as:
\begin{equation}
    \mathcal{L}_{total} = \mathcal{L}_{dif} + \alpha\mathcal{L}_{RFL},
\end{equation}
where $\alpha$ is hyperparameter to balance different losses.

It should be noticed that during the inference stage, we only adopt the AD Module to predict answers while discarding the NLR module, as the NLR module targets at providing a precise supervision for the AD module learning. The corresponding noisy distribution $\bar{y}_T$ is replaced with randomly generated Gaussian noise. The model gradually refines the random distribution through the reverse process and obtains the final predicted probability distribution $y$. The ultimate prediction class label $\hat{y}=\arg\max_l(y_l)$, where $l$ indicates the index of the categories, is the answer with the highest probability score.
\section{Experiment}
\label{sec:experiment}

\subsection{Datasets}
\label{ssec:datasets}
\textbf{VQA-RAD}~\cite{lau2018dataset} is the most widely used radiology dataset for Med-VQA, comprising 315 images and 3,515 question-answer pairs, with each image corresponding to at least one question-answer pair. $58\%$ of the questions are close-end questions and the rest are openend questions. \textbf{PathVQA}~\cite{he2020pathvqa} is constructed by extracting images along with corresponding captions from digital resources such as electronic textbooks and online libraries. It consists of 32,799 question-answer pairs, including 4,998 pathology images. Open-end questions account for $50.2\%$ of all questions and the rest are closed questions. To ensure comparability, both of the datasets have been divided following the MMQ method~\cite{do2021multiple}.
\subsection{Noise Patterns and Analysis}
\label{ssec:datasets}
Due to the lack of suitable Med-VQA datasets with noisy labels and the difficulty of creating new datasets, we adopt semantic noisy types from the natural VQA domain~\cite{zhang2023learning} to existing medical VQA datasets. ~\cite{zhang2023learning} demonstrates controlled semantic label noise by simulating human mislabeling behavior, which is more realistic than conventional random noise. We also follow the conventional VQA methods to validate our model by random label shuffling. 

The questions in Med-VQA tasks include two categories:
closed-end (e.g., Yes/No questions with one-word answers) and open-end (e.g., wh-questions with longer responses). Following ~\cite{zhang2023learning}, for closed-ended answers, we use symmetric noise by randomly flipping the answer choices. Moreover, for open-ended answers, we create a semantic space of frequent answers using a pre-trained BERT model~\cite{lee2018pre} to simulate human annotation errors. For each sample, the original answer is replaced by its closest noisy label within this semantic space. Our experiments highlight the sensitivity of prior Med-VQA methods to noisy labels, and we evaluate performance using 10\% and 20\% semantic noise ratios in the training datasets of VQA-RAD and PathVQA. Figure~\ref{show_noise} visualizes our noisy pairs and the t-SNE representation of the semantic space, where ground-truth labels are replaced with semantically similar noisy labels. Moreover, we compare the performance of different methods by introducing random noise with a 20\%  noise ratio into the training datasets of VQA-RAD and PathVQA. As shown in Table~\ref{vqa_rad} and Table~\ref{path_vqa}, at the same noise ratio, semantic noise impacts model performance more significantly than random noise and more closely resembles the type of noise encountered in actual clinical scenarios.

\subsection{Implementation Details}
Our method is trained on a single Nvidia 3090Ti GPU. We employ the Swin Transformer~\cite{liu2021swin} as the vision encoder and the BERT model~\cite{lee2018pre} as the text encoder. Our baseline model contains the Swin Transformer and BERT as encoder with transformer decoder without any answer denoise methods. The Adam optimizer is used, with an initial learning rate set to 1e-5. We adopt a batch size of 16 and a maximum of 100 epochs. The loss weight $\alpha$ is set to 0.5 for both datasets. For our model evaluation, we use only the AD Module to predict answers without the NLR module to ensure that the model parameters are at a similar level to those of other methods for comparison. The DiN framework is trained in an end-to-end fashion. The Med-VQA performance is assessed by accuracy scores.
\label{ssec:datasets}
\subsection{Performance Comparison}
\begin{figure*}[t]
\centering
\includegraphics[width=0.9\linewidth]{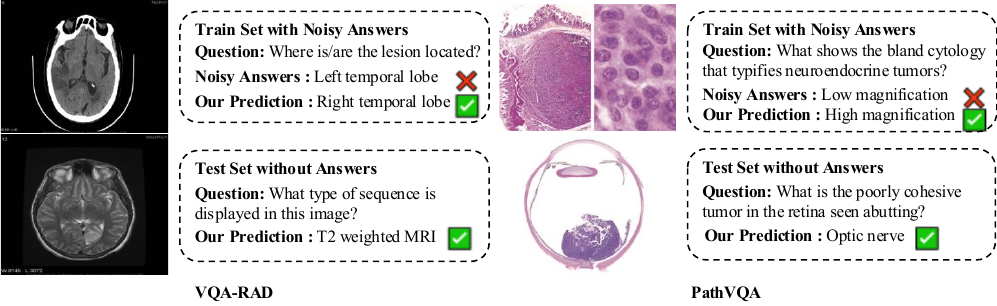}
\caption{Visualization of example results from our DiN framework on the VQA-RAD and PathVQA datasets. The first row presents examples from the training set with 10\%-Semantic Noise, where our method corrects the noisy incorrect answers during the training process. The second row shows two examples from the test sets of both datasets.}
\label{show_result}
\end{figure*}

We compare our proposed method with MMBERT~\cite{khare2021mmbert} and Q2ATransformer~\cite{liu2023q2atransformer}, which are among the top performers in Med-VQA task. Notably, previous Med-VQA methods were not designed to address the issue of noisy labels. To fill this gap, we employ two mainstream uni-modal label denoising methods to compare. SimT~\cite{guo2023handling} estimates noise transition matrices to correct label noise in a class-wise manner. CoDis~\cite{xia2023combating}, a co-teaching-based label denoising method, uses two sub-networks to select the clean label dataset based on the small-loss principle. We also compare our method with SNLC~\cite{zhang2023learning}, the first method designed for natural VQA with noisy labels. As shown in Table~\ref{vqa_rad} and Table~\ref{path_vqa}, our DiN framework consistently outperforms the compared models on both datasets. The biggest drawback of these two Med-VQA methods is that neither is designed to address the issue of noisy labels, so they cannot prevent to overfitting noisy patterns during the training process for NM-VQA tasks. 
SimT~\cite{guo2023handling} did not bring significant performance improvements, primarily due to the limitations of the noise transition matrix (NTM), which only effectively captures noise patterns in simple noisy classification tasks. CoDis~\cite{xia2023combating} performed better than NTM-based methods; but its two-branch co-teaching method doubles the parameters and uses a small-loss selection strategy better suited for simpler noise. SNLC~\cite{zhang2023learning}, designed for noisy labels in natural VQA, relies on a basic robust loss function and fails to deliver competitive performance in the medical domain. In contrast, our DiN framework is specifically tailored for noisy labels in Med-VQA tasks. Figure~\ref{show_result} visualizes four examples of predicted answers from both the training and test sets across two datasets. In the training set, our method successfully corrects semantic noisy labels post-training, while in the test set, our model generates accurate predictions in multiple instances. This design allows DiN to consistently outperform other methods, demonstrating its effectiveness in addressing the unique challenges of NM-VQA tasks.

\begin{table}[t]
\centering
\scalebox{0.7}{
\begin{tabular}{c|l|ccc}
\hline
\multirow{2}{*}{Noise Type}
&\multirow{2}{*}{Method}
&\multicolumn{3}{c}{VQA-RAD}\\\cline{3-5} 
&  & Open& Close& Overall\\ \hline
Clean&Upper Bound& 78.17& 80.08& 79.13\\ \hline
\multirow{4}{*}[-4ex]{10\%-Semantic Noise} 
&Baseline&67.86&68.85 &68.25\\
&MMBERT~\cite{khare2021mmbert}&63.36&68.21&66.42\\
&Q2ATransformer~\cite{liu2023q2atransformer}&66.79&70.17&68.48 \\
&SimT~\cite{guo2023handling}&66.93&69.73&68.04\\
&CoDis~\cite{xia2023combating}&68.02&70.54&69.53\\
&SNLC~\cite{zhang2023learning}&67.83&71.35&69.65\\
& \textbf{DiN (Ours)}&\textbf{72.68} &\textbf{75.81}&\textbf{74.24}\\ \hline

\multirow{4}{*}[-4ex]{20\%-Semantic Noise} 
&Baseline&54.23&57.14&56.01\\
&MMBERT~\cite{khare2021mmbert}&52.68&56.64&55.25\\
&Q2ATransformer~\cite{liu2023q2atransformer}&55.43&60.03&58.25\\
&SimT~\cite{guo2023handling}&53.76&57.05&55.82\\
&CoDis~\cite{xia2023combating}&54.19&57.28&56.42\\
&SNLC~\cite{zhang2023learning}&56.45&61.30&58.88\\
& \textbf{DiN (Ours)}&\textbf{58.06}&\textbf{63.17}&\textbf{62.28}\\ \hline
\multirow{4}{*}[-4ex]{20\%-Random Noise} 
&Baseline&56.71&60.30&58.88\\
&MMBERT~\cite{khare2021mmbert}&55.43&59.59&57.97\\
&Q2ATransformer~\cite{liu2023q2atransformer}&58.04&62.37&60.68\\
&SimT~\cite{guo2023handling}&57.86&62.85&60.90\\
&CoDis~\cite{xia2023combating}&59.95&60.56&60.32\\
&SNLC~\cite{zhang2023learning}&59.21&63.84&62.03\\
& \textbf{DiN (Ours)}&\textbf{61.04}&\textbf{65.77}&\textbf{63.93}\\ \hline

\end{tabular}}
\caption{Performance comparison on the VQA-RAD dataset. `Open' indicates the open-end question and `close' indicates the close-end question. The overall performance is the average performance of both question types. The best results are bold. Notably, SimT~\cite{guo2023handling} and CoDis~\cite{xia2023combating} are up-to-date representatives of NTM-based and co-teaching-based approaches, respectively.}\label{vqa_rad}
\end{table}

\begin{table}[t]
\centering
\scalebox{0.7}{
\begin{tabular}{c|l|ccc}
\hline
\multirow{2}{*}{Noise Type}
&\multirow{2}{*}{Method}
&\multicolumn{3}{c}{PathVQA}\\\cline{3-5} 
&  & Open & Close & Overall\\ \hline
Clean&Upper Bound&52.64&86.45&72.61
\\ \hline
\multirow{4}{*}[-4ex]{10\%-Semantic Noise} 
&Baseline&37.83&75.81&62.32\\
&MMBERT~\cite{khare2021mmbert}&34.18&73.21&58.73\\
&Q2ATransformer~\cite{liu2023q2atransformer}&38.42&76.85&62.61\\
&SimT~\cite{guo2023handling}&36.54&75.35& 62.06\\
&CoDis~\cite{xia2023combating}&37.86&76.28&63.02 \\
&SNLC~\cite{zhang2023learning}&38.85&78.33&64.18\\
& \textbf{DiN (Ours)}&\textbf{40.54}&\textbf{80.28}&\textbf{66.67}\\ \hline
\multirow{4}{*}[-4ex]{20\%-Semantic Noise} 
&Baseline&29.06&70.01&55.28\\
&MMBERT~\cite{khare2021mmbert}&27.13&68.35&52.41 \\
&Q2ATransformer~\cite{liu2023q2atransformer}&30.16&71.11&56.28\\
&SimT~\cite{guo2023handling}&29.38&70.69&55.93\\
&CoDis~\cite{xia2023combating}&30.36&71.18&56.55
\\
&SNLC~\cite{zhang2023learning}&30.88&71.23&56.72
\\
& \textbf{DiN (Ours)}&\textbf{32.47}&\textbf{72.56}&\textbf{58.14}\\ \hline
\multirow{4}{*}[-4ex]{20\%-Random Noise} 
&Baseline&32.29&73.53&57.03\\
&MMBERT~\cite{khare2021mmbert}&30.45&72.06&55.41\\
&Q2ATransformer~\cite{liu2023q2atransformer}&34.50&74.74&58.64\\
&SimT~\cite{guo2023handling}&35.17&74.91&59.01\\
&CoDis~\cite{xia2023combating}&35.66&74.50&58.96\\
&SNLC~\cite{zhang2023learning}&36.74&75.03&59.71\\
& \textbf{DiN (Ours)}&\textbf{37.85}&\textbf{76.16}&\textbf{60.84}\\ \hline

\end{tabular}
}
\caption{Performance comparison on the PathVQA dataset. `Open' indicates the open-end question and `close' indicates the close-end question. The overall performance is the average performance of both question types. The best results are bold. Notably, SimT~\cite{guo2023handling} and CoDis~\cite{xia2023combating} are up-to-date representatives of NTM-based and co-teaching-based approaches, respectively. 
}\label{path_vqa}
\end{table}

\begin{table}[t]
\centering
\scalebox{0.5}{
\begin{tabular}{c|cccccc|ccc|ccc}
\hline
Choice &Encoders& Classifier & ACG & AD & RFL & AA  & Open & Close & Overall& Open & Close & Overall\\ 
\hline
0&\checkmark &\checkmark &\checkmark& & & &67.86&68.85 &68.25&54.23&57.14&56.01\\
1&\checkmark &\checkmark &\checkmark&&\checkmark&&68.46&70.71&69.56&55.08&59.27&57.64\\
2&\checkmark &\checkmark & \checkmark & \checkmark & & &70.71&71.92&71.15&55.86&60.86&59.13\\
3&\checkmark &\checkmark & \checkmark & \checkmark &  &\checkmark
&70.73&73.08&72.25&56.21&61.54&60.57\\ 
4&\checkmark &\checkmark & \checkmark & \checkmark &\checkmark & &71.15&74.29&72.75&56.97&62.20&61.24\\
5&\checkmark &\checkmark &\checkmark&\checkmark&\checkmark&\checkmark&\textbf{72.68} &\textbf{75.81}& \textbf{74.24}&\textbf{58.06}&\textbf{63.17}&\textbf{62.28}\\
\hline
\end{tabular} 
}
\caption{Ablation study of our method on the VQA-RAD dataset with 10\%-Semantic Noise. ACG indicates our answer condition generator module, AD indicates our answer diffuser module. Note that our noisy label refinement module includes two parts, \textit{i.e.}, an answer adjust module (AA) and a noise robust  focal loss function (RFL). Also note that the baseline method includes the vision encoder and the text encoder (Encoder), the auxiliary classifier (Classifier) and our ACG module.}\label{ablation}
\end{table}

\begin{table}[t]
\centering

\scalebox{0.7}{
\begin{tabular}{l|l|ccc}
\hline
\multirow{2}{*}{Name}
&\multirow{2}{*}{Choice}
&\multicolumn{3}{c}{VQA-RAD with 10\% Noise}\\\cline{3-5} 
&  & Open & Close & Overall\\ \hline
\multirow{4}{*}{$L_{r}$} 
&CE~\cite{zhang2018generalized}&70.73&73.08&72.25\\
&Focal~\cite{lin2017focal}&71.43 & 73.85&73.10 \\
&SCE~\cite{wang2019symmetric}&71.86 & 73.86&73.50 \\
&RFL (ours) &\textbf{72.68} &\textbf{75.81}& \textbf{74.24}\\
\hline
\multirow{2}{*}{$L_{dif}$} 
&KL&\textbf{72.86}&75.38&74.00\\
&MSE&72.68 &\textbf{75.81}&\textbf{74.24}\\
\hline
\multirow{3}{*}{$\alpha$}
&0.1&71.79&74.23&73.75\\
&0.5&\textbf{72.68}&75.81& \textbf{74.24}\\
&1&72.16&\textbf{75.92}&74.18\\
\hline
\multirow{3}{*}{Step $T$} 
&10&72.51&75.76 &74.03\\
&50&\textbf{72.68} &\textbf{75.81}&\textbf{74.24}\\
&100&71.93&74.66& 73.87\\
\hline
\end{tabular}
}
\caption{Ablation studies on different hyperparameters, \textit{i.e.}, $\alpha$ that balances different loss functions and the number of timesteps $T$ in our answer diffuser module, and ablation studies on different options of loss functions in our noisy label refinement module.}\label{hyper}
\end{table}

\subsection{Ablation Study} 
To assess the contributions of our DiN framework, we conducted extensive ablation studies to demonstrate the efficiency of the different components of our model.

\subsubsection{Impact of Answer Condition Generator}
The ACG module provides conditional information to support the generative process in the AD module and features for training the auxiliary classifier in the NLR module. When only the encoders, ACG, and the auxliliary classifier are used for answer prediction (i.e., the fused image-question features are sent through ACG to interact with the candidate answer embeddings, and the output feature $f^c_k$ is used to train the auxiliary classifier), the accuracy is limited to $68.25\%$ overall accuracy (Choice 0 in Table~\ref{ablation}), primarily due to rapid overfitting and the inability to effectively handle noisy labels. Notably, incorporating the AD module with the conditional information from ACG significantly enhances the overall accuracy to $71.15\%$ (Choice 2). The ACG module not only enhances the auxiliary classifier's ability to capture medical domain knowledge but also strengthens the diffusion-based AD module by providing reliable Med-VQA condition information.

\subsubsection{Impact of Noisy Label Refinement}
The Noisy Label Refinement (NLR) module comprises two sub-components: Robust Loss (RFL) and answer adjustment (AA). The RFL is designed to mitigate the impact of noise on the encoders' features. As shown in Table~\ref{ablation}, incorporating the RFL leads to an increase in overall accuracy from $68.25\%$ to $69.56\%$ without using AD module and from $71.15\%$ (Choice 2) to $72.75\%$ (Choice 4) with AD module. We experimented with different combinations of robust loss functions and found that our proposed RFL is the most effective for NM-VQA tasks, as demonstrated in Table~\ref{hyper}. The answer adjustment (AA) component is specifically designed to reduce noise impact on the Answer Diffuser (AD) module. The addition of the AA component enhances the overall accuracy from $71.15\%$ (Choice 2) to $72.25\%$ (Choice 3) and further from $72.75\%$ (Choice 4) to $74.24\%$ (Choice 5) with the assistance of RFL.

\subsubsection{Impact of Answer Diffuser}
As shown in Table~\ref{ablation}, incorporating the AD module for Med-VQA prediction enhances the overall performance from $68.25\%$ (Choice 0) to $71.15\%$ (Choice 2). When all our proposed modules are combined, the overall performance further improves to $74.24\%$ (Choice 5). We also examined the hyperparameter $\alpha$, which balances the AD module and the auxiliary classifier, and the type of $L_{dif}$, as detailed in Table~\ref{hyper}. Our framework demonstrates insensitivity to both choices. We selected $\alpha=0.5$ and MSE loss as $L_{dif}$. Furthermore, we investigated the impact of varying $T$ on model performance, as presented in Table~\ref{hyper}. While increasing $T$ results in higher computational costs, our model achieves optimal performance at $T=50$. This differs from the image generation, where 1000 steps are typically used. This distinction is likely due to the precise pixel-level generation required in image tasks, whereas our model focuses on deriving a class probability distribution.

\section{Conclusion}
\label{sec:conclusion}
In conclusion, we have introduced DiN as the first approach to address the Med-VQA with noisy labels. Our method use the inherent robustness of diffusion models as AD module to deal with noisy data. We designed the ACG module to fuse image, question features and candidate answering embeddings, generating condition information supporting both the AD and NLR modules. Additionally, the NLR module reduces the effect of noisy labels with robust loss functions and refines the original answers to produce pseudo-labels, further improving model performance. Extensive experiments validate the superiority of our DiN framework. While our method has some limitations, the diffusion model introduces Gaussian noise during training, which may slightly increase training time but does not affect inference time.
\small \bibliographystyle{ieeenat_fullname} \bibliography{main}


\end{document}